\documentclass[sigconf,natbib=false]{acmart}
\AtBeginDocument{%
  \providecommand\BibTeX{{%
    Bib\TeX}}}
\usepackage{comment}
\usepackage[capitalize]{cleveref}
\usepackage{tikz}
\usepackage{tabularx}
\usepackage{algorithm}
\usepackage{algorithmic}

\copyrightyear{2026}
\acmYear{2026}
\setcopyright{cc}
\setcctype{by}
\acmConference[SAC '26]{The 41st ACM/SIGAPP Symposium on Applied Computing}{March 23--27, 2026}{Thessaloniki, Greece}
\acmBooktitle{The 41st ACM/SIGAPP Symposium on Applied Computing (SAC '26), March 23--27, 2026, Thessaloniki, Greece}
\acmPrice{}
\acmDOI{10.1145/3748522.3779772}
\acmISBN{979-8-4007-2294-3/2026/03}

\RequirePackage[
  datamodel=acmdatamodel,
  style=acmnumeric,
  ]{biblatex}

\title{Iterative In-Context Learning to Enhance LLMs Abstract Reasoning: The Case-Study of Algebraic Tasks}

\renewcommand{\shorttitle}{In-Context Learning for LLMs’ Abstract Reasoning}

\newtheorem{example}{Example}

\author[S.~Fioravanti]{Stefano Fioravanti}
\affiliation{%
  \institution{University of Padua and Charles University Prague}
  \city{Padua}
  \country{Italy/Prague,Czachia}
}

\email{stefano.fioravanti@unipd.it}

\author[M.~Zavatteri]{Matteo Zavatteri}
\affiliation{%
  \institution{University of Padua}
  \city{Padua}
  \country{Italy}
}
\email{matteo.zavatteri@unipd.it}

\author[R.~Confalonieri]{Roberto Confalonieri}
\affiliation{%
  \institution{University of Padua}
  \city{Padua}
  \country{Italy}
}
\email{roberto.confalonieri@unipd.it}

\author[K.~Zeinalipour]{Kamyar Zeinalipour}
\affiliation{%
  \institution{University of Siena}
  \city{Siena}
  \country{Italy}
}
\email{kamyar.zeinalipour2@unisi.it}

\author[P.~Frazzetto]{Paolo Frazzetto}
\affiliation{%
  \institution{University of Padua}
  \city{Padua}
  \country{Italy}
}
\email{paolo.frazzetto@unipd.it}

\author[A.~Sperduti]{Alessandro Sperduti}
\affiliation{%
  \institution{University of Padua}
  \city{Padua}
  \country{Italy}
}
\email{alessandro.sperduti@unipd.it}

\author[N.~Navarin]{Nicol\`o Navarin}
\affiliation{%
  \institution{University of Padua}
  \city{Padua}
  \country{Italy}
}
\email{nicolo.navarin@unipd.it}

\begin{document}

\begin{abstract}
LLMs face significant challenges in systematic generalization, particularly when dealing with reasoning tasks requiring compositional rules and handling out-of-distribution examples. To address these challenges, we introduce an in-context learning methodology that improves the generalization capabilities of general purpose LLMs. Our approach employs an iterative example selection strategy, which incrementally constructs a tailored set of few-shot examples optimized to enhance model's performance on a given task. As a proof of concept, we apply this methodology to the resolution of algebraic expressions involving non-standard simplification rules, according to which the priority of addition and multiplication is changed.  
Our findings indicate that LLMs exhibit limited proficiency in these mathematical tasks. We further demonstrate that LLMs reasoning benefits from our iterative shot selection prompting strategy integrated with explicit reasoning instructions. Crucially, our experiments reveal that some LLMs achieve better generalization performances when prompted with simpler few-shot examples rather than complex ones following the test data distribution. 
\end{abstract}


\maketitle

\section{Introduction}



Large Language Models (LLMs) have shown remarkable capabilities in understanding and generating human-like content. They achieve impressive results in core NLP tasks such as machine translation, summarization, sentiment analysis, and question answering \cite{brown2020language,openai2023gpt4}. These models are capable of few-shot and even zero-shot learning, significantly reducing the need for task-specific supervision \cite{chowdhery2022palm}. Moreover, their ability to model complex linguistic patterns enables strong performance across diverse domains. 

Despite some promising achievements \cite{Lample2019DeepLF}, recent research has shown that
state-of-the-art transformer architectures and LLMs still lack systematic and compositional generalization skills~\cite{shojaee2025,Saxton2019AnalysingMR}. 
Furthermore, 
LLMs perform well when test data resembles training examples but fail to generalize to patterns beyond their observed data. 
These findings underscore the inherent limitations of LLMs in  compositional tasks and extrapolating beyond their training data. 


To mitigate these problems,  
we propose a few-shot in-context learning
methodology aimed at improving the generalization capabilities of LLMs in solving  algorithmic tasks. Our methodology  selects the shots 
by querying iteratively the LLMs to solve an algorithmic task specified in a given dataset. To this end,  
it adds to the prompt the examples that were wrongly computed, 
 employing  contemporary approaches to enhance LLMs performances like chain-of-thoughts \cite{diao-etal-2024-active, wang2022self, DBLP:conf/nips/Wei0SBIXCLZ22}. Our methodology ensures that the few-shot examples become more representative and diverse over time, thereby improving the  model’s capacity to generalize. This prompting approach  mirrors the natural human learning process---learning through failure followed by targeted feedback---similar to curriculum learning~\cite{10.1145/1553374.1553380}. 

As a case study for our methodology we chose the domain of mathematics. 
In particular, 
we focus on algebraic  
formula simplification 
for evaluating the systematic generalization capabilities of LLMs.  
Simplifying nested formulas involving operations on lists of integers, as well as arithmetic and algebraic expressions, presents a significant challenge \cite{Petruzzellis2024-mi}. 
Formula simplification demands the precise and consistent application of transformation rules, making it an ideal testbed to provide a rigorous assessment of the model's ability to learn and apply algorithmic procedures. 

Unlike previous
studies that primarily evaluate LLMs on basic mathematical problems~\cite{gairmathabel,luo2023wizardmath,yu2023metamath} or propose math-specific LLMs ~\cite{yang2024qwen25,xin2024deepseekp}, our investigation focuses on vanilla LLMs' reasoning  ability to solve algebraic expressions involving non-standard operators according to which the priority of multiplication and addition is inverted. 
In this way, we can test the LLM reasoning abilities on a task that probably was not encountered in its training phase. In this simple task, which  can also be performed by low-level high school students, we show the limited capabilities of LLMs to perform out-of-distribution tasks and to reason about mathematical formulas. 
We evaluate 
the performance of LLMs in formula simplification tasks by constructing a set of synthetic datasets with increasing levels of difficulty. Through extensive experiments 
we show LLMs' limited capability of reasoning through mathematical tasks.
We further show 
that few-shot prompting can mitigate this issue and that our methodology 
can even further enhance the performances of a given LLM using the same number of shots. 


The contributions of this paper are the following:
\begin{enumerate}
    \item We provide a novel strategy to enhance LLMs' reasoning capabilities via few-shot prompting. This is done in two steps: (i) we introduce an iterative, error-driven approach for synthesizing a compact set of in-context examples (shots), which can be seen as a training phase for the prompt; (ii) we use the synthesized shots to solve the underlying reasoning task in a standard few-shot prompting setup.
    \item We construct 5 synthetic datasets of increasing difficulty for algebraic expressions with a non-standard order of operations, and we show that modern LLMs exhibit profound limitations on this kind of out-of-distribution reasoning task.
    \item We release all datasets, prompts, and scripts used in our experiments to support reproducibility and further research.\footnote{\texttt{https://github.com/StefanoF66/llms-math}}
\end{enumerate}

\section{Related work}\label{sec:related}

\textbf{Large Language Models (LLMs).}
Recent advances in prompt engineering have shown that the reasoning capabilities of large language models can be significantly enhanced through the use of targeted prompting techniques \cite{DBLP:conf/iclr/0002WSLCNCZ23,DBLP:conf/nips/Wei0SBIXCLZ22}. Chain-of-Thought (CoT) prompting, in particular, has been effective in leveraging the autoregressive nature of these models to elicit multi-step reasoning, thereby improving performance on complex tasks \cite{diao-etal-2024-active, wang2022self, DBLP:conf/nips/Wei0SBIXCLZ22}. 
LLMs have shown notable success in various symbolic reasoning tasks, such as coin flip prediction, last-letter concatenation \cite{DBLP:conf/nips/Wei0SBIXCLZ22}, and Boolean variable assignment \cite{DBLP:conf/nips/AnilWALMRSGDN22}. These tasks bear a close resemblance to those examined in the present work, as they involve synthetically generated instances that can be resolved using relatively simple algorithmic procedures. Nevertheless, despite these advances, LLMs continue to exhibit limitations in problems that require robust systematic generalization~\cite{press-etal-2023-measuring}. Finally, the work in~\cite{diao-etal-2024-active} proposes an active prompting strategy to adapt LLMs
to different tasks with task-specific example prompts. Compared with it, our approach does not require any annotations and it relies on an iterative methodology.  

\textbf{Systematic Generalization.}
A substantial body of research has examined the extent to which neural networks are capable of learning in a systematic and compositional manner \cite{hupkes2020compositionality,testolin2023can}. To evaluate these abilities, several benchmarks have been introduced. Notably, ListOps \cite{DBLP:conf/naacl/NangiaB18} and CTL \cite{DBLP:journals/corr/abs-1802-06467} serve as prototypical examples of formula simplification tasks, where models must compute intermediate results—either implicitly or explicitly—in order to evaluate a given expression. 
While these datasets include standard expressions, in this paper we exploit non-standard order of operations to assess the model's reasoning abilities in tasks that were not seen during training. 
Additional benchmarks in this domain include symbolic integration and differentiation tasks \cite{Lample2019DeepLF}, as well as polynomial simplification challenges \cite{agarwal2021analyzing}. Modern neural architectures demonstrate varying degrees of success in systematic and compositional generalization \cite{DBLP:conf/iclr/CsordasIS22,Lake2023-qb,DBLP:conf/acl/OntanonAFC22}. Recent studies have highlighted the potential of modular designs, such as Mixture of Experts (MoE) models, which exhibit emergent specialization in expert modules and show improved performance in both in-distribution and out-of-distribution settings \cite{DBLP:conf/nips/MittalBL22}. These findings support the hypothesis that modularity may play a critical role in fostering generalization beyond the training distribution.

\textbf{Rewriting Systems and Neural Networks.}
The integration of symbolic rewriting paradigms into neural models has been explored in a variety of forms. Early work in unsupervised settings proposed neural encodings where network weights correspond to symbolic tokens subject to rewriting \cite{icnc09}. Subsequent approaches have combined handcrafted feature engineering with feed-forward architectures to address algebraic simplification tasks \cite{Cai2018-xx}. More recently, reinforcement learning was employed to design adaptive rewriting systems capable of learning to identify sub-expressions for transformation and selecting suitable rewrite rules \cite{DBLP:conf/nips/ChenT19}. More specifically, standard algebraic problems were also treated in \cite{ai4ua,Keskin2023156} and \cite{Petruzzellis2024ANR} where the authors developed a modular architecture designed to learn a general procedure for solving nested mathematical formulas by only relying on a minimal set of training examples.
In this paper, we consider the reasoning abilities of general purpose LLMs, that is, we do not make any assumptions on the internal structure of the models. 

\textbf{LLMs and mathematical reasoning.} Driven by the rapid development of LLMs since 2021, the number of math-specific LLMs (Math-LLMs) has grown steadily, alongside enhanced support for multilingual and multi-modal capabilities. The landscape was marked by the introduction of models like GPT-f \cite{polu2020generative} and Minerva \cite{lewkowycz2022solving}, with Hypertree Proof Search \cite{lample2022hypertree} and Jiuzhang 1.0 \cite{zhao2022jiuzhang} highlighting advancements in theorem proving and mathematical question understanding capabilities, respectively. In 2023 we witnessed a surge in diversity and specialization, alongside multi-modal support from models like Skywork-Math \cite{zeng2024skywork}. In 2024, there was a clear focus on enhancing mathematical instruction (\textit{e.g.}, Qwen2.5-Math \cite{yang2024qwen25}) and proof (\textit{e.g.}, DeepSeek-Proof \cite{xin2024deepseekp}) capabilities. That year also witnessed the emergence of Math-LLMs with a vision component, such as MathGLM-Vision \cite{yang2024mathglm}. 
Studying purpose-built LLMs is beyond the scope of this paper, as our focus is on the general reasoning capabilities of LLMs. Evaluating the proposed approach's significance by comparing it with specialized LLMs is left as future work.

\section{A Novel Few-Shot Prompting Methodology}\label{sec:methodology}







Prompting is the practice of providing input (a prompt) to an LLM to guide the LLM itself through the generation of a desired output. Prompts can be as simple as questions or as complex as structured instructions with examples. Prompting allows for asking questions, generating creative content, summarizing, transforming or processing text, and solving problems or complete tasks. Prompting is foundational to how LLMs interact with users in their natural language. Few-shot prompting is a specific prompting technique where the prompt encloses a few examples of the task the LLM is expected to solve, followed by a new previously unseen input related to the same task. Prompting can be divided into:
\begin{itemize}
	\item zero-shot: no examples are given, only the instructions;
	\item one-shot: one example is provided;
	\item few-shot: a few examples are provided.
\end{itemize}

However, a critical point of all these approaches remains: 
\emph{How do we choose these shots?}    
In what follows, we use the terms \emph{shot} and \emph{example} interchangeably. We prefer ``example'' in the textual exposition for readability, but we keep ``shot'' in figures and tables to stay consistent with the few-shot prompting terminology.

We now provide a general method for few-shot prompting that consists of two phases:
\begin{enumerate}
    \item {\bf Few-shot synthesis}: We provide an iterative approach to synthesize a set of shots to be used later in the prompt. This phase can be seen as a way to train the prompt;
    \item {\bf Few-shot prompting evaluation}: We prompt the LLM by using the shots synthesized in the first phase to solve the underlying tasks. 
\end{enumerate}

\begin{figure}[t]
    \centering
    \begin{tikzpicture}

    \node[rectangle,draw,thick, rounded corners, minimum height=80pt, minimum width=50pt] (LLM) {LLM};
    \node[rectangle,draw,thick, rounded corners, minimum width=140pt,minimum height=80pt] (PO) [right of=LLM,xshift=120pt,label={Prompt Agent}] {};

    \draw[->,draw,thick] ([yshift=20pt]PO.south west) -- node[auto,sloped] {prompt}  ([yshift=20pt]LLM.south east);

    \draw[->,draw,thick] ([yshift=-20pt]LLM.north east) -- node[auto,sloped] {answer}  ([yshift=-20pt]PO.north west);

    \node[rectangle,draw,thick, rounded corners, minimum width=40pt,minimum height=20pt] (ANALYZE) [right of=LLM,xshift=80pt,yshift=20pt,align=center] {Answer\\analyzer};

    \node[rectangle,draw,thick, rounded corners, minimum width=40pt,minimum height=20pt] (GENERATESHOTS) [right of=LLM,xshift=160pt,yshift=20pt,align=center] {Shot\\generator};

    \node[rectangle,draw,thick, rounded corners, minimum width=40pt,minimum height=20pt] (SHOTS) [below of=GENERATESHOTS,yshift=-15pt,align=center] {Shot\\set};

    \node[rectangle,draw,thick, rounded corners, minimum width=40pt,minimum height=20pt] (GENERATEPROMPT) [below of=ANALYZE,yshift=-15pt,align=center] {Prompt\\generator};

    \draw[->,draw,dotted] ([yshift=-20pt]PO.north west) -- node[auto,sloped] {} (ANALYZE);

    \draw[->,draw,dotted] (ANALYZE) -- node[auto] {\emph{wrong}} (GENERATESHOTS);
    
    \draw[->,draw,dotted] ([xshift=-10pt]ANALYZE.south) -- node[auto,right] {\emph{correct}} ([xshift=-10pt]GENERATEPROMPT.north);

    \draw[->,draw,dotted] ([xshift=10pt]GENERATESHOTS.south) -- node[auto,left] {\emph{new data}}  ([xshift=10pt]SHOTS.north);

    \draw[->,draw,dotted] (SHOTS) -- node[above] {\emph{shots}}  (GENERATEPROMPT);

     \draw[->,draw,dotted] (GENERATEPROMPT) -- node[auto,sloped] {} ([yshift=20pt]PO.south west);
    \end{tikzpicture}
    \caption{Iterative few-shot synthesis.}
    \label{fig.shot-selection}
\end{figure}
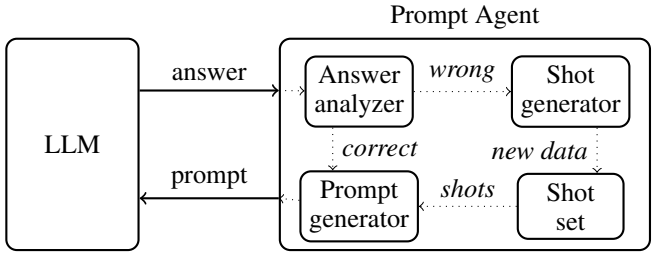

Figure~\ref{fig.shot-selection} shows a high-level view of the architecture for the first phase (few-shot synthesis). The architecture is general,  iterative, and intended to function for a broad number of underlying tasks.  The most external level of the architecture consists of two interacting agents:
\begin{enumerate}
    \item The {\bf Prompt Agent}: an intelligent agent in charge of interacting with the LLM over time;
    \item The {\bf LLM}: a black-box agent.
\end{enumerate}


Communication between these two agents happens iteratively in a Q\&A fashion. In our instantiation, the Prompt Agent processes a finite \emph{calibration} dataset $\mathcal{D}_{\text{cal}}$ one instance at a time: at each iteration it builds a prompt for a single expression from $\mathcal{D}_{\text{cal}}$, receives the model's answer, and possibly updates the shot set. The loop terminates after all instances in $\mathcal{D}_{\text{cal}}$ have been processed, so the maximum number of iterations is $|\mathcal{D}_{\text{cal}}|$.

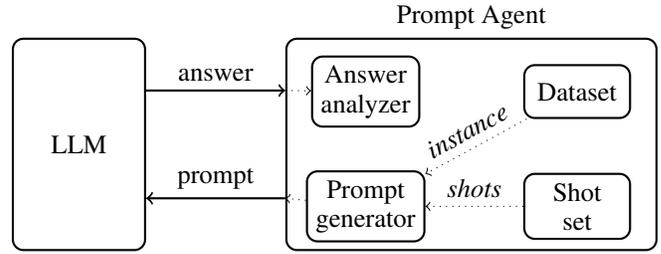
\begin{figure}[t]
    \centering
    \begin{tikzpicture}

    \node[rectangle,draw,thick, rounded corners, minimum height=80pt, minimum width=50pt] (LLM) {LLM};
    \node[rectangle,draw,thick, rounded corners, minimum width=140pt,minimum height=80pt] (PO) [right of=LLM,xshift=120pt,label={Prompt Agent}] {};

    \draw[->,draw,thick] ([yshift=20pt]PO.south west) -- node[auto,sloped] {prompt}  ([yshift=20pt]LLM.south east);

    \draw[->,draw,thick] ([yshift=-20pt]LLM.north east) -- node[auto,sloped] {answer}  ([yshift=-20pt]PO.north west);

    \node[rectangle,draw,thick, rounded corners, minimum width=40pt,minimum height=20pt] (ANALYZE) [right of=LLM,xshift=80pt,yshift=20pt,align=center] {Answer\\analyzer};

     \node[rectangle,draw,thick, rounded corners, minimum width=40pt,minimum height=20pt] (DATASET) [right of=LLM,xshift=160pt,yshift=20pt,align=center] {Dataset};

    \node[rectangle,draw,thick, rounded corners, minimum width=40pt,minimum height=20pt] (SHOTS) [below of=GENERATESHOTS,yshift=-15pt,align=center] {Shot\\set};

    \node[rectangle,draw,thick, rounded corners, minimum width=40pt,minimum height=20pt] (GENERATEPROMPT) [below of=ANALYZE,yshift=-15pt,align=center] {Prompt\\generator};

    \draw[->,draw,dotted] (DATASET) -- node[auto,sloped] {\emph{instance}} (GENERATEPROMPT);

    \draw[->,draw,dotted] ([yshift=-20pt]PO.north west) -- node[auto,sloped] {} (ANALYZE);

     \draw[->,draw,dotted] (SHOTS) -- node[above] {\emph{shots}}  (GENERATEPROMPT);
    
     \draw[->,draw,dotted] (GENERATEPROMPT) -- node[auto,sloped] {} ([yshift=20pt]PO.south west);
    \end{tikzpicture}
    \caption{Few-shot prompting evaluation.}
    \label{fig.few-shot}

\end{figure}

 The generic iteration of the shot synthesis process is as follows. The Prompt Agent generates a prompt and asks the LLM to solve some specific task. This step may use previous computed shots and/or information coming from external data sources. The LLM replies to the Prompt Agent with an answer. At this point the Prompt Agent analyzes the answer produced by the LLM and two cases are possible:
\begin{enumerate}
    \item The LLM provided the correct answer. In this case no shot is generated;
    \item The LLM provided the wrong answer. In this case the Prompt Agent generates new shots from that answer and adds them to the shot set.
\end{enumerate}
Regardless of the case, the approach continues with the next iteration (possibly considering an augmented set of shots). Eventually, the set of shots contains the shots that can be used to run the LLM on the test set.

This feedback loop ensures that the few-shot examples become more representative and diverse over time, thus enhancing the model’s capacity to generalize. More importantly, we constrain the number of total examples in the test prompt to remain fixed, focusing on quality rather than quantity. 
In particular, this prompting approach mirrors the natural human learning process similar to curriculum learning~\cite{10.1145/1553374.1553380}.
Just as students gradually refine their understanding by revisiting examples they previously found challenging, our iterative synthesis mechanism emphasizes examples that highlight the model's current weaknesses. This alignment with human-like incremental learning intuitively allows the model to develop a more nuanced internal representation of the problem space. By continually integrating mistakes into the learning context and correcting them with explicit reasoning chains, the model simulates a form of guided practice akin to pedagogical tutoring. Over time, this leads to a more diverse and representative shot synthesis, covering edge cases and rare expression structures that static prompting might overlook. Moreover, by maintaining a fixed prompt length, the strategy ensures that learning is directed toward content quality rather than sheer volume, paralleling how human learners prioritize clarity and conceptual depth over rote memorization. In essence, our method bridges the gap between static few-shot prompting and dynamic curriculum learning~\cite{10.1145/1553374.1553380}, 
offering a principled framework for strengthening LLMs’ generalization in mathematical reasoning tasks. Clearly, the exact implementation of the Prompt Agent and its internal modules is left to the designer with specific domain knowledge.

Finally, Figure~\ref{fig.few-shot} provides a schema of our few-shot prompting for the second phase. In this phase the Prompt Agent generates a prompt starting from an instance of the underlying task contained in a dataset. Such a prompt is augmented with the set of shots synthesized in phase 1. In this phase, we only care about evaluating the answer of the LLM to estimate how good the shot synthesis phase was. Clearly, assuming a finite dataset and assuming that we process (in order) one instance at a time, such an evaluation phase eventually terminates.


\paragraph{Instantiation of the Prompt Agent.}
In our algebraic case study, the Answer Analyzer is implemented as a simple exact evaluator: given an input expression and the non-standard priority rule, it computes the ground-truth result symbolically and compares it with the model's answer. Whenever a mismatch is detected, the Shot Generator creates a new example consisting of the original expression together with a step-by-step computation under the modified semantics (see Example~\ref{ex1}). These corrected computations are then reused as demonstrations in subsequent prompts. In other words, the model receives explicit chain-of-thought style examples of how to solve instances it previously failed on.\\
This design assumes the availability of a domain-specific agent that is \emph{proficient} in the target task. For many practically relevant domains, such an agent can be implemented using existing tools (e.g., symbolic solvers for algebra, program interpreters, or rule-based systems). In settings where only approximate or noisy solvers are available, our methodology can still be applied, but the quality of the synthesized shots will depend on the reliability of the underlying agent. A systematic analysis of this trade-off is left for future work.

\section{The Case Study of Algebraic Tasks}
\label{sec:case_study}


To evaluate the capacity of LLMs to perform systematic mathematical reasoning, we design a series of experiments focused on algebraic formula simplification. We deviate from standard mathematical conventions by introducing a modified operator precedence rule: \textbf{addition (\texttt{+}) is evaluated before multiplication (\texttt{*})}. This simple yet cognitively challenging setup enables us to probe whether LLMs can generalize rule-based transformations beyond surface-level pattern matching. In our case study, we instantiated the few-shot synthesis presented in the previous section as follows.
\begin{itemize}
    \item As an external source of data we rely on a synthetic dataset of (input, output) pairs; 
    \item At the beginning the set of shots is empty;
    \item At every iteration, the prompt generator picks a single  expression from the dataset following the order of appearance plus all the shots selected so far in order to generate a specific prompt to solve the expression according to the given semantics (see below);

    \item If the LLM fails on some prompt,  the shot generator generates a new shot that coincides with the original (input, correct output) instance from the dataset and adds such a shot to the set of selected shots.
\end{itemize}


As previously mentioned, in our case study we
alter the order of operations. In standard arithmetic, multiplication takes precedence over addition. However, in our tasks, we instruct the model to treat addition as the higher-priority operator. For example, given the expression \texttt{3 + 2 * 4}, the expected answer under modified rules is \texttt{3 + 2 = 5}, then \texttt{5 * 4 = 20}, not the conventional \texttt{11}.

This reversal operator precedence requires the model to:
\begin{itemize}
    \item override learned mathematical priors from pretraining;
    \item apply transformation rules consistently;
    \item perform intermediate computations step by step.
\end{itemize}

Such a scenario forms a rigorous benchmark for systematic generalization, particularly in out-of-distribution settings.

\subsection{Dataset Generation}

To investigate the capability of LLMs to perform out-of-distribution mathematical tasks we design an algorithm to create synthetic datasets of well-formed random expressions (Algorithm~\ref{alg:expression_generation}). 

The algorithm implements a recursive procedure designed to generate randomized mathematical expressions with controlled complexity and structural diversity involving an input set of 
numbers $N$ and a set of operations $O$. The approach recursively builds expressions by selecting operations and operands at each recursion step. Starting from an initial call at a depth $d=0$ (where depth is interpreted as the nesting level of brackets), the function continues recursively, increasing $d$ until reaching a maximum depth $\mathit{max\_depth}$. 

The complexity of generated expressions is controlled by the parameter $comp > 0$. 
This parameter directly affects the probability distribution over the number of operands  in a subexpression at depth $k$. Specifically, the probabilities are defined as follows:  
$p_4 = \min(1, \frac{1}{\mathit{comp}})$, $p_3 = (1-p_4)*\min(1, \frac{2}{\mathit{comp}})$,  $p_2 = (1-p_4-p_3)*\min(1,  \frac{\lceil comp*0.3\rceil}{\mathit{comp}})$, $p_1 = (1-p_4-p_3-p_2)$, when $d \neq 0$. For the case $d=0$, we enforce the constraints $p_2 = (1-p_4-p_3)$ and $p_1 = 0$ to guarantee at least one non-trivial expression (i.e., involving more than a single operator, also including repetitions of the same operator) at depth $1$.  
Intuitively, increasing $\mathit{comp}$ reduces the probability of generating subexpressions with more operands, thus lowering the overall structural complexity of the expression. 

\begin{algorithm}[t]\label{alg:dataset}
\caption{$\mathtt{GenerateExpression}(\mathit{max\_depth}, \mathit{comp})$}
\small
\label{alg:expression_generation}
\begin{algorithmic}[1]
\REQUIRE $d \gets 0$, Number of elements $N$, set of operations $O$
\IF{$d == \mathit{max\_depth}$}
    \RETURN Random choice from set of elements $N$
\ENDIF
\STATE Compute $[p_1, p_2, p_3, p_4]$ based on $\mathit{\mathit{comp}}$

\STATE Sample $p \sim U(0,1)$
\IF{$p \leq p_4$}
    \STATE Select four random operations $\mathit{op}_1, \mathit{op}_2, \mathit{op}_3, \mathit{op}_4$ from $O$ 
    \STATE Recursive generation of $E_1, E_2, E_3, E_4, E_5$ with $d+1$  
    \STATE 
    $E \gets (E_1 \ \mathit{op}_1 \ E_2 \ \mathit{op}_2 \ E_3 \ \mathit{op}_3 \ E_4 \ \mathit{op}_4 \ E_5)$
\ELSIF{$p \leq p_3 + p_4$}
    \STATE Select three random operations $\mathit{op}_1, \mathit{op}_2, \mathit{op}_3$ from $O$
    \STATE Recursive  generation of  $E_1, E_2, E_3, E_4$ with $d+1$ 
    \STATE 
    $E \gets (E_1 \ \mathit{op}_1 \ E_2 \ \mathit{op}_2 \ E_3 \ \mathit{op}_3 \ E_4)$
\ELSIF{$p \leq p_2 + p_3 + p_4$}
    \STATE Select two random operations $\mathit{op}_1, \mathit{op}_2$ from $O$
    \STATE Recursive  generation of $E_1, E_2, E_3$ with $d+1$ 
    \STATE 
    $E \gets (E_1 \ \mathit{op}_1 \ E_2 \ \mathit{op}_2 \ E_3)$
\ELSE
    \STATE Select one random operation $\mathit{op}_1$
    \STATE Recursive  generation of $E_1, E_2$ with $d+1$ 
    \STATE 
    $E \gets (E_1 \ \mathit{op}_1 \ E_2)$
\ENDIF
\RETURN $E$ \hfill \textit{(final generated expression)}
\end{algorithmic}
\end{algorithm}
The algorithm is deliberately designed to control the complexity of generated expressions, ensuring that the created datasets  exhibit uniform levels of difficulty. By standardizing complexity across instances, the evaluation focuses solely on the reasoning capabilities of the models, eliminating potential biases stemming from variations in problem structure.

Furthermore, controlling expression complexity enables the systematic creation of difficulty tiers (e.g., easy, medium, and hard). This is crucial for assessing how well models generalize at increasing levels of reasoning depth. 


We use two measures of complexity of the generated expression:
\begin{enumerate}
    \item the {\em depth of nesting}, i.e., the number of nested brackets;
    \item the {\em complexity of subexpressions}\footnote{By ``subexpression'' we mean any expression enclosed within a pair of brackets.}.
\end{enumerate}

\paragraph{Depth of the nesting.}
Increasing the depth of the expression generally increases its difficulty. For example:
\begin{itemize}
    \item \textbf{Depth 1 (Easy)}: $(3 + 2*  1)$
    \item \textbf{Depth 2 (Medium)}: $(3 + (5 * 2) + 4)$
    \item \textbf{Depth 3 (Hard)}: $(4 + (2 * (9 + 3))) + 5)$
\end{itemize}

\paragraph{Complexity of  subexpressions.} 
The number of operators inside each subexpression of the expression tree also contributes to overall difficulty. 
For example:
\begin{itemize}
    \item \textbf{One operator (Easy)}: $2 *  3$
    \item \textbf{Two operators (Medium)}: $2 + 4 * 6$
    \item \textbf{Three operators (Hard)}: $2 + 4 * 8 + 5$
\end{itemize}

\paragraph{Synthetic datasets.} 
For the experimental evaluation,  we generated $5$ synthetic datasets, each containing $200$ datapoints. The datasets are parametrized by $(\mathit{depth}, \mathit{comp})$, where $\mathit{depth}$ denotes the maximum number of bracket levels, and $\mathit{comp}$ controls the probability of generating subexpressions with more than one operator. The datasets are defined as \texttt{db(depth, comp)} $\in \{(1,6), (2,20),$ $(2,10), (2,6), (3,20)\}$.   

\subsection{Prompting Strategies}
\label{sec.prompting-strategies}
Three primary prompting strategies can be considered as repair strategies, namely zero-shot, few-shot, and chain-of-thought. Each of these aims at guiding LLMs through the reasoning process with various levels injection of external information.

\subsubsection{Few-shot prompting}

Few-shot prompting supplements the task with a small number of worked examples that adhere to the same rule change. In particular, in the experimental part we have varied the number of shots from $0$ to $50$. These examples act as in-context demonstrations, helping the model infer the transformation logic.

\begin{example}
\label{ex1}
Prompt Example:

Q: Given the following examples:
Computations example1:

Expression: ((3 * 8 + 1) + (2 + 8) + (2 + 5))

Steps: [8 + 1 = 9, 3 * 9 = 27, 2 + 8 = 10, 27 + 10 = 37, 2 + 5 = 7, 37 + 7 = 44, 44]

Computations example2:
Expression: ((4 + 0) * (3 + 1) + (2 + 4))
Steps: [4 + 0 = 4, 3 + 1 = 4, 2 + 4 = 6, 4 + 6 = 10, 4 * 10 = 40, 40]

Simplify the following expression ((3*5+4) + (2 + 9 * 0))  where + has priority over *. 

\end{example}

Few-shot prompting helps disambiguate the task format and serves as an implicit training mechanism. It also provides insight into the LLM’s ability to extrapolate from limited examples.

We further add a short system prompt containing some basic information about the task and basic examples to force a structured answer. Furthermore, in the experimental setting we study two different setups for prompt and shots, providing the step in a list format (\texttt{PV1}), as in the example, or as line separated text (\texttt{PV2}). 

\subsubsection{Chain-of-Thought prompting}

Following recent advances in the literature --- see, for example, \cite{Petruzzellis2024ANR} --- we extensively adopt Chain-of-Thought (CoT) prompting as a core component of our methodology. We explicitly instruct the model to reason step by step. CoT reasoning was proven effective in previous works involving arithmetic and logic problems. Here, we combine it with the few-shot setup for maximum effectiveness by adding ``\verb|A: Let's think step by step.|'' at the end of the prompt as in \cite{Petruzzellis2024ANR}.

\section{Experimental Results}\label{sec:experiments}

We conducted comprehensive experiments using five newly developed synthetic datasets to assess the mathematical reasoning capabilities of contemporary LLMs in term of accuracy over the final result of a formula. Specifically, we evaluated two 
variants of Gemini \cite{deepmind2023gemini}, namely \texttt{gemini-2.0-flash-exp} (GMN2.0) and \texttt{gemini-2.0-flash-thinking-exp-01-21} (GMN2.0-R), as well as two versions of DeepSeek \cite{bi2024deepseek}, \texttt{deepseek-chat} (DS-C) and \texttt{deepseek} \texttt{-reasoner} (DS-R). For GMN2.0, GMN2.0-R, and DS-C. We report on the results averaged over three independent runs to account for variability. Due to the significant computational and financial costs associated with DS-R, its evaluation was restricted to  a single run conducted per experimental setting. The complete set of experiments required approximately two months to execute.

In addition to our core methodology, we conducted a prompt engineering study by designing and evaluating two distinct prompt variants tailored to our case study. These prompt formulations, referred to as \texttt{PV1} and \texttt{PV2}, were selected based on preliminary experimentation and employed throughout our evaluation. Notably, we observed significant discrepancies in model behavior depending on the prompt used, underscoring the impact of prompt design on model performance and for this reason we included both prompts in our experimental study. 

In this work we address the following research questions (RQs):

\begin{description}
    \item[RQ1:] How does the number of in-context examples (shots) affect the performance of LLMs on non-standard algebraic reasoning tasks?
    \item[RQ2:] How does the \emph{selection} of shots (random vs. error-driven vs. out-of-distribution easy shots) and the prompt format influence the resulting performance?
    \item[RQ3:] How do these effects interact with dataset complexity and model family (reasoning vs. non-reasoning LLMs)?
\end{description}

These RQs directly correspond to our contributions: RQ1 and RQ2 focus on the proposed iterative prompting methodology, while RQ3 analyzes its robustness across datasets and models.

\subsection{Research Q1: Impact of the Shot Number}

\begin{figure}[t]
\vspace{-10pt}

    \centering
    \includegraphics[width=0.5\textwidth]{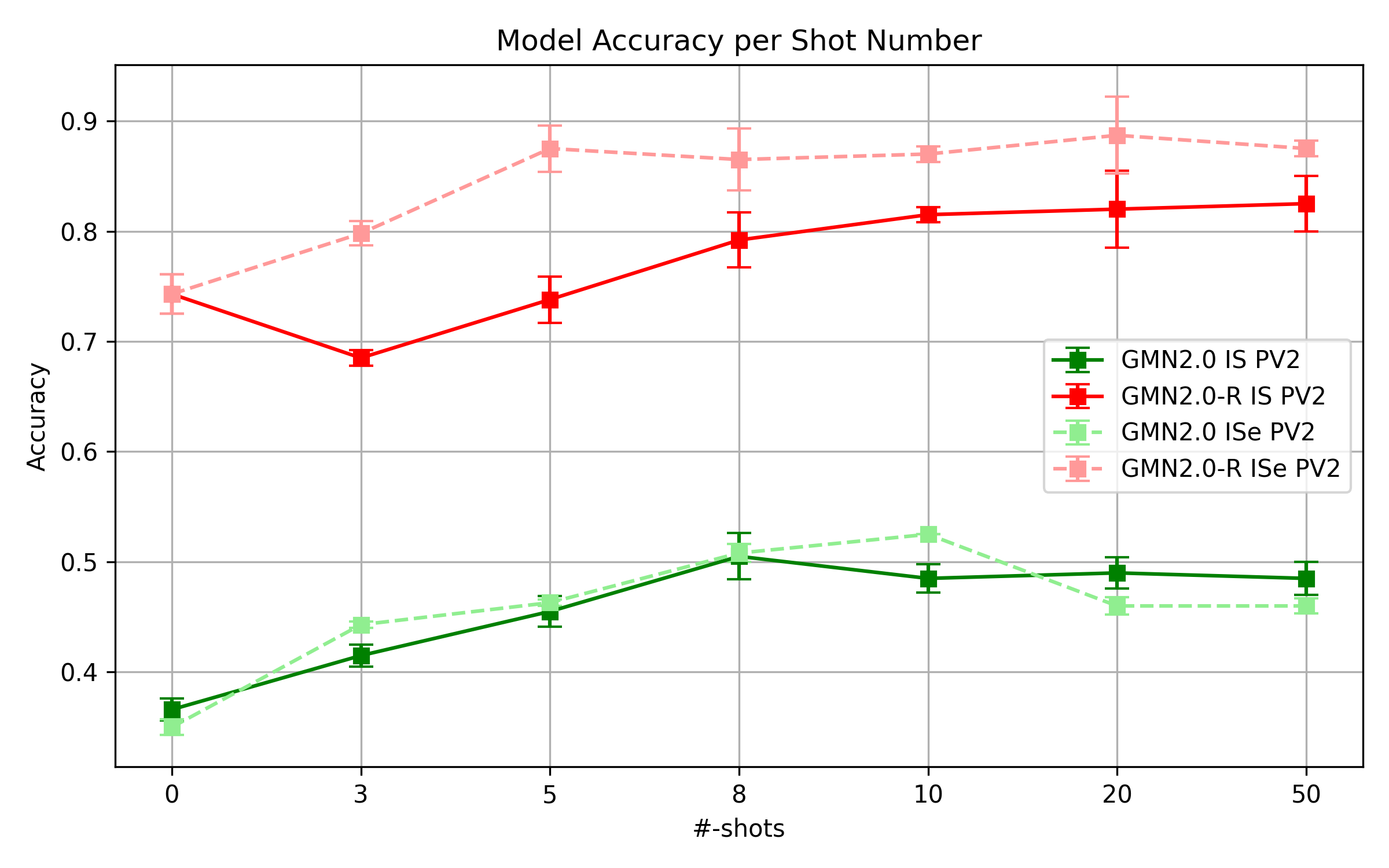} 
    \caption{ Change of performances over number of shots of the models GMN2.0 and GMN2.0-R using the prompt \texttt{PV2} over the dataset \texttt{db(2,20)}.  In the table we use --IS-- when we employ our iterative shot selection approach, --e-- stands for shot selected from a dataset of parameters \texttt{(1,6)}.}
    \label{table1}
    \vspace{-2pt}
\end{figure}

The first study we present examines the impact of the number of shots included in the prompts on the performance of the selected LLMs. Figure~\ref{table1} presents a comparative analysis concerning how the performances of the models GMN2.0 and GMN2.0-R change based on the number of shots used, evaluated on the dataset \texttt{db(2,20)}. Specifically, the prompt type \texttt{PV2} was chosen for this evaluation, as it consistently yielded the highest overall performance in the 0-shot setting with the Gemini models. Our findings indicate that GMN2.0-R, which incorporates an explicit reasoning module, consistently outperforms GMN2.0 across prompting strategies. 


We further observed that both models generally benefit from few-shot prompting in out-of-distribution generalization tasks when using \texttt{PV2}. This is evidenced by the performance gains achieved through our iterative shot selection procedure (see \cref{fig.shot-selection}) utilizing examples drawn from a dataset sharing the distribution with \texttt{db(1,6)} (denoted as ISe in \cref{table1}). 

Although only partially illustrated in Figure~\ref{table1}, we observed a consistent trend across experiments: model performance tends to stabilize around the 10-shot mark, with little to no improvement—and in most cases a decline—when the number of shots exceeds 50. This saturation effect suggests a diminishing return in few-shot learning beyond a certain threshold, likely due to increased length of the prompt and cognitive or representational overload for the model. Notably, this behavior was consistently replicated across all datasets, prompt types, and shot types in our evaluation, underscoring the generality and robustness of this pattern, and thereby justifying our decision to fix the number of shots to $10$ in subsequent research investigations.

\subsection{Research Q2: Impact of the Shot and Prompt Selection}

\begin{figure}[t]
\vspace{-10pt}

    \centering
    \includegraphics[width=0.5\textwidth]{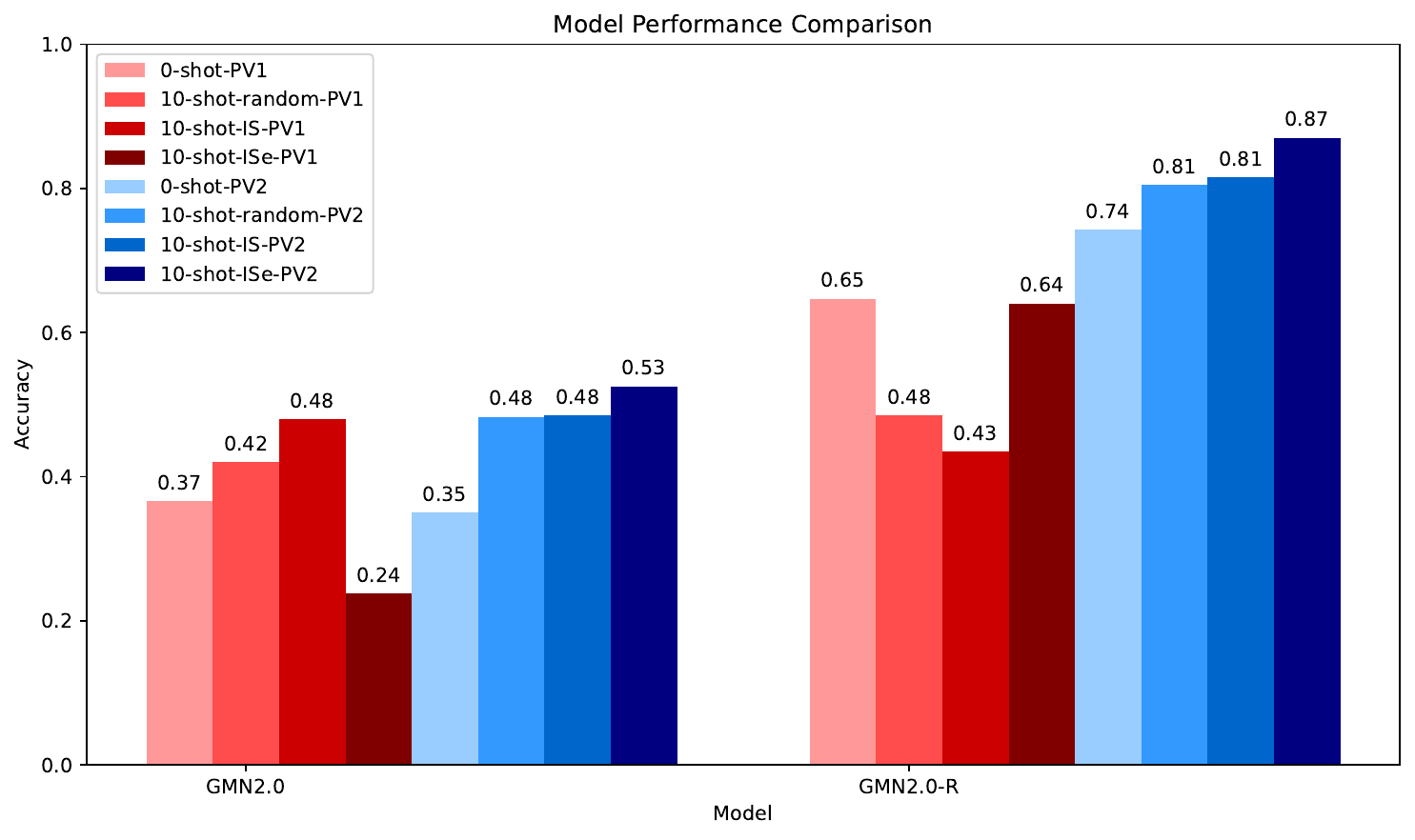} 
    \caption{ Performances in terms of accuracy over changing of type of shots for the models GMN2.0 and GMN2.0-R using the prompts \texttt{PV1} and \texttt{PV2} over the dataset \texttt{db(2,20)}. In the table we use --IS-- when we employ our iterative shot selection approach, --e-- stands for shot selected from a dataset of parameters \texttt{(1,6)} while --random-- is used for randomly selected shots within the same distribution of the test set.}
    \label{fig:2}
    \vspace{-3pt}

\end{figure}

We observed that the prompt type and shot selection can strongly affect LLM performance.
Figure~\ref{fig:2} presents a detailed comparative analysis of the models GMN2.0 and GMN2.0-R on the dataset \texttt{db(2,20)}, with a focus on the effects of prompt type and shot selection strategies. We examine the performances under two prompting configurations, \texttt{PV1} and \texttt{PV2} with particular emphasis on \texttt{PV2}, which demonstrated the highest overall performance in the 0-shot setting for Gemini models. The number of shots is fixed at 10, a choice motivated by empirical observations indicating that model performance generally saturates with 10 shots (c.f., Figure~\ref{table1}), as we have seen from the previous study.

We evaluated three shot selection strategies: (i) random—shots randomly sampled from the same distribution as the test set, (ii) (IS)—shots selected using our iterative shot selection method within the test distribution, and (iii) (ISe)—shots selected via the same iterative approach but drawn from a simpler, out-of-distribution dataset (\texttt{db(1,6)}).
The results indicate that GMN2.0 exhibits relatively stable performance across all shot selection methods. Specifically,
for prompt \texttt{PV1}, the highest accuracy is achieved using the IS strategy and we can see a degradation of performances using shots ISe. For prompt
\texttt{PV2}
the 10-shot strategy consistently improves over the 0-shot, and the ISe configuration yields the best results. This suggests that GMN2.0 is capable of leveraging structurally simpler examples to improve generalization out-of-distribution on more complex tasks.

GMN2.0-R, the version of the model equipped with an explicit reasoning module, displays improved results over GMN2.0, but also greater sensitivity to both prompt and shot selection.
Notably, when using \texttt{PV1} the performance are not affected much or are deteriorated by the inclusion of shots, independently from the technique used for shot selection.
With prompt \texttt{PV2}, we can see that all prompting technique increase the model accuracy.
However, the performance improves markedly when ISe shots are introduced, indicating that this model particularly benefits from training signals derived from simpler examples to improve generalization out-of-distribution on more complex tasks.

Importantly, the patterns described here are not confined to a single dataset; similar trends were consistently observed across all datasets considered in our study, highlighting the robustness of these findings.

\begin{table*}[ht]
\caption{Experimental results in terms of accuracy for all two Gemini models across the datasets comparing 0-shot (0-shot), 10 random shots (10-shots), 10-iteratively selected shots (10-shot IS) within the same distribution of the test set, and 10-iteratively selected shots (10-shot ISe) from a dataset of parameters (1,6) with the prompt version obtaining overall the best performances in 0-shot.}
\label{tab:exp_best}
\resizebox{\textwidth}{!}{
\begin{minipage}{\textwidth}
\setlength{\tabcolsep}{4pt}
\begin{tabularx}{\textwidth}{@{}l c *{8}{>{\centering\arraybackslash}X}@{}}
\toprule[1.5pt]
\multicolumn{2}{c}{} & 
\multicolumn{4}{c}{GMN2.0 \texttt{PV2}} & 
\multicolumn{4}{c}{GMN2.0-R \texttt{PV2}} \\

\cmidrule(lr){3-6} \cmidrule(lr){7-10}

\multicolumn{2}{c}{Dataset} & 
0-shot & 10-shot & 10-shot IS & 10-shot ISe & 
0-shot & 10-shot & 10-shot IS & 10-shot ISe \\

\midrule[1.5pt]
& \texttt{db(2,20)} 
& $0.350 \pm 0.007$ & $0.483 \pm 0.012$ &  $0.485 \pm 0.013$ & $\mathbf{0.525} \pm 0.0$
& $0.743 \pm 0.18$ & $0.805 \pm 0.011$ 
& $0.815 \pm 0.007$ & $\mathbf{0.87} \pm 0.018$\\

\midrule
& \texttt{db(2,10)} 
& $0.217 \pm 0.004$ &  $0.271 \pm 0.003$ & $0.281 \pm 0.08$ & $\mathbf{0.315} \pm 0.01$ 
& $0.565 \pm 0.28$ & $0.71 \pm 0.021$ 
& $0.725 \pm 0.0$ & $\mathbf{0.785} \pm 0.01$\\

\midrule
& \texttt{db(2,6)} 
& $0.133 \pm 0.003$ & $0.17 \pm 0.07$ & $0.166 \pm 0.06$ & $\mathbf{0.190} \pm 0.01$ 
& $0.465 \pm 0.008$ & $0.675 \pm 0.007$ 
& $0.678 \pm 0.011$ & $\mathbf{0.765} \pm 0.03$\\

\midrule
& \texttt{db(1,6)} 
& $0.635 \pm 0.005$ & $0.905 \pm 0.0$ & $\mathbf{0.967} \pm 0.01$ & $\mathbf{0.967} \pm 0.01$ & $0.947 \pm 0.018$ 
& $\mathbf{0.975}\pm0.011$ & $0.972 \pm 0.012$ & $0.972 \pm 0.012$ \\

\midrule
& \texttt{db(3,20)} 
& $0.145 \pm 0.005$ & $0.137 \pm 0.015$ & $0.138 \pm 0.06$ & $\mathbf{0.146} \pm 0.02$ 
& $0.445 \pm 0.03$ 
& $0.45 \pm 0.007$ & $0.465 \pm 0.011$ & $\mathbf{0.585} \pm 0.05$ \\

\bottomrule[1.5pt]
\end{tabularx}
\end{minipage}
}
\end{table*}

   \begin{table*}[ht]
    \vspace{-5pt}
\caption{Experimental results in terms of accuracy for all two DeepSeek models across the five datasets comparing 0-shot (0-shot), 10 random shots (10-shots), 10-iteratively selected shots (10-shot IS) within the same distribution of the test set, and 10-iteratively selected shots (10-shot ISe) from a dataset of parameters (1,6) with the prompt version obtaining overall the best performances in 0-shot.}
\label{tab:exp_best_2}
\resizebox{\textwidth}{!}{
\begin{minipage}{\textwidth}
\setlength{\tabcolsep}{4pt}
\begin{tabularx}{\textwidth}{@{}l c *{8}{>{\centering\arraybackslash}X}@{}}
\toprule[1.5pt]
\multicolumn{2}{c}{} & 
\multicolumn{4}{c}{DS-C \texttt{PV1}} & 
\multicolumn{4}{c}{DS-R \texttt{PV1}} \\

\cmidrule(lr){3-6} \cmidrule(lr){7-10}

\multicolumn{2}{c}{Dataset} & 
0-shot & 10-shot & 10-shot IS & 10-shot ISe & 
0-shot & 10-shot & 10-shot IS & 10-shot ISe \\

\midrule[1.5pt]
& \texttt{db(2,20)} 
& $0.338 \pm 0.008$ & $0.390 \pm 0.009$ & $0.413 \pm 0.02$ & $\mathbf{0.435} \pm 0.02$ 
& $0.675$ & $0.665$ & $0.79$ & $\mathbf{0.87}$ \\

\midrule
& \texttt{db(2,10)} 
& $0.21 \pm 0.01$ & $0.277 \pm 0.01$ & $0.288 \pm 0.02$ & $\mathbf{0.33} \pm 0.011$
& $0.55$ & $0.595$ & $0.625$ & $\mathbf{0.68}$ \\

\midrule
& \texttt{db(2,6)} 
& $0.148 \pm 0.003$ & $0.167 \pm 0.08$ & $0.168 \pm 0.01$ & $\mathbf{0.185} \pm 0.01$
& $0.36$ & $0.485$ & $0.545$ & $\mathbf{0.61}$ \\

\midrule
& \texttt{db(1,6)} 
& $0.705 \pm 0.022$ & $0.77 \pm 0.01$ & $0.833 \pm 0.01$ & $\mathbf{0.845} \pm 0.01$
& $0.955$ & $0.805$ & $0.94$ & $\textbf{0.96}$ \\

\midrule
& \texttt{db(3,20)} 
& $0.175 \pm 0.005$ & $0.213 \pm 0.003$ & $\mathbf{0.216} \pm 0.01$ & $0.214 \pm 0.008$
& $0.315$ & $0.355$ & $0.32$ & $\mathbf{0.47}$ \\

\bottomrule[1.5pt]
\end{tabularx}
\end{minipage}
}
\end{table*}

\subsection{Research Q3: Overall performances}

In this final study, we conduct a comprehensive analysis across the synthetic datasets, prompts, type of shots, and considered models. 

Tables~\ref{tab:exp_best} and~\ref{tab:exp_best_2} present the performance results of the four evaluated models—GMN2.0, GMN2.0-R, DS-C, and DS-R—across the studied five synthetic datasets using prompt versions \texttt{PV2} and \texttt{PV1}, respectively. The choice of the prompt versions is based on their superior or equivalent overall accuracy in the baseline 0-shot.

The results span various prompting conditions, including 0-shot, standard 10-shot, and iteratively selected shots (10-shot IS and 10-shot ISe), providing a comprehensive comparison across prompting strategies. 
%
The tables clearly show that dataset complexity, as governed by the structural parameters, significantly impacts model performance (with respect to both the parameters used): as complexity increases, performance consistently declines across all models and prompting configurations. This trend supports the intended difficulty scaling of the synthetic datasets and highlights the challenges models face in more intricate reasoning tasks.

Consistent behavioral patterns emerge across the prompt versions that reinforce the insights from Figure~\ref{fig:2}. GMN2.0 (\cref{tab:exp_best}) demonstrates strong robustness to prompt variation, with similar baseline (0-shot) results for both \texttt{PV1} and \texttt{PV2}. Moreover, it benefits steadily from the addition of few-shot examples, with peak performance consistently achieved through our iterative shot selection method (IS). The overall best accuracy is reached with prompt \texttt{PV2} and (ISe) shots. This suggests that GMN2.0 generalizes well from easy 
examples to more complex test cases 
when using \texttt{PV2}.

GMN2.0-R (\cref{tab:exp_best}) shows a similar behavior but with a slightly greater sensitivity to prompt structure, probably caused by the reasoning module. Under \texttt{PV2},  the model demonstrates substantial gains from few-shot prompting—especially when using ISe examples—achieving some of the highest accuracy values observed. The results highlight the model's dependency on particular prompts to leverage its reasoning capabilities effectively. We note that performance does not improve on the easiest  \texttt{db(1,6)}. This can be attributed to the already high baseline 0-shot accuracy. In this setting, miscomputed expressions constitute rare outliers, and few-shot prompting proves less effectiveness, as the provided examples are unlikely to capture such infrequent cases.

DS-C (\cref{tab:exp_best_2}) shows also a performance profile more aligned with GMN2.0, demonstrating robustness to prompt types in baseline settings. Interestingly, however, DS-C attains its best baseline results with under \texttt{PV1}, diverging from the trends observed in the Gemini models, which offer their best performances with \texttt{PV2}. This might suggest differences in how each architecture incorporates context or handles inductive biases introduced by few-shot examples.  DS-R (\cref{tab:exp_best_2}) exhibits behavior comparable to that of DS-C. In this case as well, the inclusion of a reasoning module consistently enhances performance across all settings.


Overall, our findings confirm that the considered task presents a significant challenge even for modern LLMs. However, our iterative shot selection consistently enhances performance across models, validating its value as a lightweight yet powerful strategy. Notably, models with integrated reasoning modules (e.g., GMN2.0-R) often achieve superior peak performance, but simpler base models like GMN2.0 can close this gap through informed shot selection—(as shown by GMN2.0 over \texttt{db(1,6)}, where the IS strategy allows the model to achieve comparable performance to GMN2.0R) offering a competitive and computationally efficient alternative in few-shot scenarios, especially on relatively simple reasoning tasks.

\subsection{Reproducibility}\label{sec:reproducibility}

For each dataset, we fix the train--calibration--test splits and reuse them across all models and prompting conditions. The synthetic expression generator (Algorithm~\ref{alg:expression_generation}), together with the exact parameters $(\texttt{depth}, \texttt{comp})$ for each dataset, is included in the supplementary material. We also release:

\begin{itemize}
    \item the full list of prompts, including both prompt variants (\texttt{PV1}, \texttt{PV2}), and the system instructions used to elicit step-by-step reasoning;
    \item the final shot sets produced by the iterative selection procedure for each model/dataset combination;
    \item the scripts used to query the models and compute accuracy.
\end{itemize}

These resources will be made publicly available upon publication to facilitate exact reproduction and extension of our results.

\section{Conclusions and Future Works}\label{sec:conclusion}

This paper explores the systematic generalization reasoning capabilities of LLMs and proposes a novel iterative few-shot repair methodology that improves the abstract reasoning of general purpose LLMs. The methodology is applied to non-standard algebraic tasks using a set of 5 synthetic datasets that introduce non-standard orders of operations and varying levels of complexity. The study uncovers notable limitations in LLMs' ability to handle novel mathematical tasks, revealing that their reasoning is often brittle when faced with unfamiliar problem structures. 
The proposed 
methodology significantly improves LLMs' reasoning performance by choosing informative examples over failures. Our study also finds that LLMs tend to perform better when provided with easier examples, showing interesting out-of-distribution shot generalization capabilities. These results highlight the need for improved training paradigms that better support LLMs' generalization capabilities. 
Our work represents an important step toward understanding the limitations of LLMs in abstract reasoning 
and provides a foundation for future research aimed at developing more 
robust LLMs-based AI systems.

While our work addresses critical limitations and proposes an initial mitigation strategy of LLMs in abstract mathematical tasks, it marks only the first step toward a broader vision: building language model-based assistants for mathematical and scientific discoveries. Achieving this goal will require several future research directions.

\textbf{Handling Custom Algebraic Structures.} We could expand beyond our current synthetic arithmetic setting to more complex mathematical expressions, including symbolic identities, polynomials, matrices, modular arithmetic. This would better test LLMs' compositionality and transfer learning across mathematical domains.

\textbf{Incorporating Soft Constraints through Fine-Tuning.} Fine-tuning models with soft constraints, rewarding logical execution and penalizing invalid operations, can help internalize the problem structure more deeply than prompting alone. Techniques like contrastive or reinforcement-based fine-tuning are promising paths to explore.


\textbf{Extending to Higher-Level Mathematical Tasks.}  Moving from expression evaluation to higher-level tasks—such as equation solving,  proof generation/verification—will be essential. These tasks demand logical rigor and structured reasoning, ultimately bringing us closer to the vision of LLMs as true collaborators in mathematical and scientific discoveries.

\paragraph{Limitations and future work.}
Our study is limited in several ways. First, we focus on a single family of tasks, namely synthetic algebraic expressions with non-standard operator precedence. While this setup is useful to probe systematic generalization, it does not cover other important reasoning domains such as logic puzzles, program synthesis, or real-world word problems. Extending our methodology to qualitatively different tasks is an important direction for future work. Second, our experimental evaluation is restricted to two model families (Gemini and DeepSeek), primarily due to computational and monetary constraints. Evaluating the proposed iterative in-context learning strategy on popular open-weight models such as LLaMA, Qwen or Gemma would provide a more complete picture of its generality. Finally, although we compare several prompting strategies (zero-shot, few-shot, CoT, random vs. iterative selection), we do not exhaust all possible baselines (e.g., contrastive few-shot prompting or more sophisticated zero-CoT variants). Our results should therefore be seen as a first step toward understanding iterative shot synthesis, rather than a definitive comparison of prompting techniques.

\begin{acks}
This work was supported by: the National Recovery and Resilience Plan of Italian Ministry of University and Research funded by the European Union – NextGenerationEU, Mission 4 Component 2 Investment 1.5 - Call for tender No. 3277 of 30 December 2021, project code: ECS00000043, Concession Decree No. 1058 of June 23, 2022, CUP C43C22000340006, project title ``iNEST: Interconnected Nord-Est Innovation Ecosystem''; Mission 4 Component 2 Investment 1.3 - Call for tender No. 341 of March 15, 2022, project code PE0000013, Concession Decree No. 1555 of October 11, 2022, CUP C63C22000770006, project title ``Future AI Research (FAIR) - Spoke 2 Integrative AI - Symbolic conditioning of Graph Generative Models (SymboliG)''; and by the Charles University grant number PRIMUS/24/SCI/008.
\end{acks}




\printbibliography
\end{document}